\documentclass[lettersize,journal]{IEEEtran}
\usepackage{amsmath,amsfonts}
\usepackage{algorithmic}
\usepackage{algorithm}
\usepackage{array}
\usepackage[caption=false,font=normalsize,labelfont=sf,textfont=sf]{subfig}
\usepackage{textcomp}
\usepackage{stfloats}
\usepackage{url}
\usepackage{verbatim}
\usepackage{graphicx}
\usepackage{cite}
\usepackage{pgfplots}
\usetikzlibrary{arrows.meta}

\begin{document}

\title{Non-monotonic causal discovery with\\ Kolmogorov-Arnold Fuzzy Cognitive Maps}

\author{Jose L. Salmeron
\thanks{Prof. Salmeron is with the School of Engineering, CUNEF Universidad, Madrid, Spain.}
\thanks{Please cite as: J.L. Salmeron, "Non-monotonic causal discovery with Kolmogorov-Arnold Fuzzy Cognitive Maps," in IEEE Transactions on Fuzzy Systems, forthcoming, doi: 10.1109/TFUZZ.2026.3680681.}
}



\markboth{Salmeron: Non-Monotonic Causal Discovery with Kolmogorov-Arnold FCM}%
{Salmeron: Non-Monotonic Causal Discovery with Kolmogorov-Arnold FCM}


\maketitle

\begin{abstract}
Fuzzy Cognitive Maps constitute a neuro-symbolic paradigm for modeling complex dynamic systems, widely adopted for their inherent interpretability and recurrent inference capabilities. However, the standard FCM formulation, characterized by scalar synaptic weights and monotonic activation functions, is fundamentally constrained in modeling non-monotonic causal dependencies, thereby limiting its efficacy in systems governed by saturation effects or periodic dynamics. To overcome this topological restriction, this research proposes the Kolmogorov-Arnold Fuzzy Cognitive Map (KA-FCM), a novel architecture that redefines the causal transmission mechanism. Drawing upon the Kolmogorov-Arnold representation theorem, static scalar weights are replaced with learnable, univariate B-spline functions located on the model edges. This fundamental modification shifts the non-linearity from the nodes' aggregation phase directly to the causal influence phase. This modification allows for the modeling of arbitrary, non-monotonic causal relationships without increasing the graph density or introducing hidden layers. The proposed architecture is validated against both baselines (standard FCM trained with Particle Swarm Optimization) and universal black-box approximators (Multi-Layer Perceptron) across three distinct domains: non-monotonic inference (Yerkes-Dodson law), symbolic regression, and chaotic time-series forecasting. Experimental results demonstrate that KA-FCMs significantly outperform conventional architectures and achieve competitive accuracy relative to MLPs, while preserving graph-based interpretability and enabling the explicit extraction of mathematical laws from the learned edges.
\end{abstract}


\section{Introduction}

\IEEEPARstart{T}{he} modeling of complex dynamic systems requires a trade-off between predictive accuracy and semantic interpretability. Fuzzy Cognitive Maps (FCMs) \cite{kosko1986fuzzy} have bridged this gap by representing knowledge as a signed directed graph where nodes denote concepts and edges represent causal influence. Unlike black-box approaches such as Deep Neural Networks, FCMs allow domain experts to directly inspect the adjacency matrix. This structural accessibility ensures the interpretability of the underlying inference mechanism. This property has driven their adoption across diverse high-stakes domains, ranging from medical decision support \cite{salmeron2017medical} and environmental policy planning to industrial process control \cite{salmeron2017learning}.

Classical FCMs are constrained by the theoretical limitation of the weighted linearity assumption. Typically, the influence of one concept upon another is modeled as the product of the activation value and a static scalar weight. This approach restricts causality to strict monotonicity. Under this assumption, the model is limited to representing effects that increase or decrease in direct proportion to their causes. While sufficient for elementary regulatory systems, this simplification fails to capture complex phenomena characterized by thresholds, optima, saturation, or periodicity. For instance, in biological systems, the relationship between enzyme concentration and reaction rate is rarely linear, characterized by a saturation curve. Similarly, in psychology, the Yerkes-Dodson law dictates that performance enhances with physiological or mental arousal only up to a critical point, beyond which it deteriorates. A standard FCM, constrained by scalar weights, is unable of representing this non-monotonic dynamic without the introduction of hidden nodes that reduce interpretability.

To mitigate these shortcomings, research efforts have shifted toward the development and implementation of High-Order and Deep FCMs \cite{wang2020deep}. These variations typically stack multiple layers of concepts (nodes) or introduce complex non-linearities at the node level. While these methods successfully augment predictive performance in time-series forecasting, they frequently obfuscate the direct causal link between source and target nodes. The interpretability of causal edges is severely compromised when interactions are processed through multiple hidden transformations. Consequently, the FCM evolves toward a black-box architecture functionally equivalent to a Multi-Layer Perceptron (MLP). As a result, the literature often reflects a trade-off between the high performance of deep learning and the transparency inherent in cognitive maps.

Recently, Liu et al. \cite{liu2024kan} proposed Kolmogorov-Arnold Networks (KANs) as a structural alternative to the conventional Multi-Layer Perceptron. Derived from the Kolmogorov-Arnold representation theorem \cite{kolmogorov1957representation}, KANs allocate the non-linearity along the edges instead of the traditional node-based activation approach. This research proposes that the KAN architecture serves as a extension for Fuzzy Cognitive Maps. The integration of learnable univariate functions onto the causal edges leads to the development of the Kolmogorov-Arnold Fuzzy Cognitive Map (KA-FCM). This novel architecture enables an edge to learn complex behaviors, such as a concept enhancing another up to a threshold before inhibiting it. In this sense, the model effectively captures non-monotonicity within a single, observable connection.

The main contributions of this research are as follows:
\begin{itemize}
    \item A theoretical framework for Kolmogorov-Arnold Fuzzy Cognitive Maps that redefines the adjacency matrix as a functional adjacency matrix. In this architecture, each entry is replaced by a learnable spline, thereby extending the mathematical foundations of the paradigm.
    \item A framework that extends the expressivity of FCMs through learnable functions while retaining their symbolic interpretability. This ensures that the causal transparency is preserved, addressing a critical limitation found in existing Deep FCM methodologies.
    \item The empirical validation of the KA-FCM's capacity to perform symbolic regression across causal edges, demonstrating the model's ability to extract explicit functional relationships from observed data, while enhancing the transparency of the discovered dynamics.
    \item A benchmark on non-linear function approximation and chaotic time-series forecasting demonstrating that KA-FCM significantly outperforms traditional approaches in terms of reconstruction error.
\end{itemize}

The remainder of this paper is organized as follows. Section \ref{sec:related_works} provides a review of the related works about Fuzzy Cognitive Maps and Kolmogorov-Arnold theory. Section \ref{sec:method} presents the methodological proposal, the B-spline edge parameterization, and the learning algorithm of the Kolmogorov-Arnold Fuzzy Cognitive Map (KA-FCM). Section \ref{sec:exps} describes the experimental framework and analyzes the results obtained across three case studies: non-monotonic inference, symbolic regression, and chaotic time-series forecasting. Finally, Section \ref{sec:conclusion} summarizes the findings and provides concluding remarks.

\section{Related works}
\label{sec:related_works}


Since Kosko's seminal proposal \cite{kosko1986fuzzy}, significant research efforts have focused on three main axes: the development of learning algorithms for weight estimation \cite{salmeron2024blind, salmeron2025concurrent}, the creation of structural extensions for temporal or granular precision \cite{napoles2021construction, vanhoenshoven2020pseudoinverse, salmeron2019learning, salmeron2019uncertainty}, and the incorporation of deep learning principles \cite{napoles2025inverse}.

For the purpose of this paper, this section is going to focus on learning algorithms, high-order and granular extensions, explainability and deep architectures.

\subsection{Learning algorithms}
The estimation of the causal weight matrix $W$ is the central challenge in the design of Fuzzy Cognitive Maps. Early methodologies relied heavily on expert knowledge, a process often constrained by subjective bias and the inherent complexity of manually quantifying causal strengths in large-scale systems \cite{wu2020fast}. Data-driven methods emerged to address these concerns, incorporating biological synaptic plasticity as an early framework for learning. To adjust weights based on concept activation correlations, NHL and AHL were introduced \cite{papageorgiou2004learning}. However, despite their computational efficiency, these approaches present known constraints in achieving global convergence, particularly when dealing with complex, non-linear relationships \cite{salmeron2019uncertainty}.

To overcome these limitations, the field transitioned toward global optimization strategies, leveraging Evolutionary Computation and Swarm Intelligence to estimate adjacency matrices from historical data. Particle Swarm Optimization (PSO) and its variants have emerged as a standard for training FCMs, demonstrating superior convergence properties in complex domains such as time-series forecasting \cite{salmeron2016dynamic} or decision support systems \cite{salmeron2017medical}. Hybrid metaheuristics have also been developed to enhance the balance between exploration and exploitation. Notable contributions include Memetic Algorithms, which integrate local refinement procedures within a global evolutionary search \cite{salmeron2017learning}, and Modified Asexual Reproduction Optimization algorithms achieving faster convergence in high-dimensional spaces \cite{salmeron2019learning}.

These population-based methods successfully minimize the error between the concept state vector and historical data by navigating the global search space. However, a fundamental structural limitation persists across all these optimization paradigms. Whether trained via dynamic PSO, Memetic Algorithms, or Asexual Reproduction, the objective remains the optimization of static scalar weights. This formulation strictly assumes that interactions between concepts are linear and monotonic. As a result, static scalar weights are inherently insufficient for modeling complex behaviors like saturation or non-monotonicity. In such cases, the causal polarity is not constant but depends dynamically on the activation level of the source concept \cite{vanhoenshoven2020pseudoinverse}.

\subsection{High-Order, granular and grey extensions}
High-Order FCMs (HO-FCMs) emerged as a solution to the limitations of first-order models, incorporating historical states to capture more complex dynamics \cite{nikseresht2025time}. By incorporating dependencies on multiple previous time steps ($t-1, t-2, \dots, t-k$), HO-FCMs can model long-term temporal dependencies more effectively than standard architectures \cite{high_order_2023}. While this architecture enhances forecasting accuracy in multivariate time series, it significantly increases the dimensionality of the weight matrix. This expansion often complicates the interpretation of immediate causal drivers by distributing the causal influence across multiple time lags.

Parallel to temporal extensions, efforts to mitigate epistemic uncertainty and data sparsity have led to the integration of Granular Computing and Grey Systems Theory into the cognitive mapping framework. The Fuzzy Grey Cognitive Map (FGCM) \cite{salmeron2010modelling} further addresses these challenges by employing grey numbers to represent intervals with unknown distributions, thereby facilitating the modeling of uncertain systems where only small datasets are available \cite{salmeron2015fgcm}. FGCMs have demonstrated robustness in reliability engineering and process control by applying interval-valued operations to the inference mechanism \cite{froelich2014evolutionary, salmeron2014fuzzy, salmeron2016autonomous}. Also, granular FCMs \cite{granular_framework_2022, granular_extension_2023} replace scalar activation values with information granules, allowing the model to process non-numeric or imprecise linguistic information.  

However, despite their improved ability to handle vagueness and noise, both Granular FCMs and FGCMs typically rely on interval arithmetic or defuzzification processes. These operations preserve the monotonicity of the underlying causal relations. Consequently, these extensions remain constrained in their ability to model complex, non-monotonic functional dependencies directly on the network edges.

\subsection{Explainability and deep architectures}
The trade-off between approximation capacity and model interpretability constitutes a fundamental challenge in computational intelligence. Recent comparative studies \cite{CHO2025128781, explainability_lamda_2023, guerrero2020lrp} suggest that preserving an explicit causal graph structure is essential for aligning FCMs (and computational intelligence methods) with Explainable Artificial Intelligence (XAI). 

The need for enhanced non-linear approximation in standard cognitive maps has led to the development of Deep Fuzzy Cognitive Map architectures \cite{LI2024111771}. These approaches incorporate hidden layers between input and output concepts to model highly non-linear functions. Deep architectures frequently achieve superior accuracy through the introduction of hidden concepts. However, their lack of semantic interpretation compromises the direct causal logic that distinguishes FCMs from black-box models. 

Concurrently, other lines of research have focused on analyzing the limit state space of quasi-nonlinear FCMs to assess stability \cite{quasi_nonlinear_limit_2023, napoles2025inverse}. However, these models primarily study the dynamics of a given topology, failing to capture the underlying functional forms that govern the interactions between concepts.

\section{Methodological proposal}
\label{sec:method}

\subsection{Problem formulation}

Consider a discrete-time dynamic system composed of $N$ conceptual nodes, denoted as a set $\mathcal{C} = \{c_1, c_2, \dots, c_N\}$. The activation state of each specific concept $c_i$ at time step $t$ is represented by a scalar value $c_i(t) \in [0,1]$ \cite{kosko1986fuzzy}. Consequently, the overall system state is denoted by the vector $c(t)=[c_{1}(t),c_{2}(t),...,c_{N}(t)]^{T}\in[0,1]^{N}$ [1]. Traditional First-Order FCMs rely on a static adjacency matrix $W\in\mathbb{R}^{N\times N}$ to represent causal inference. Here, each scalar entry $w_{ij}$ defines the strength of the relationship between a source concept $c_j$ and a target concept $c_i$ \cite{napoles2025inverse}.

\begin{equation}
    c_i(t+1) = f \left( \sum_{j=1}^{N} w_{ij} c_j(t) \right)
\end{equation}

\noindent where $f: \mathbb{R} \to [0,1]$ is a monotonic activation function, usually unipolar sigmoid or hyperbolic tangent \cite{bueno2009benchmarking}. 

This approach limits causal dependencies to monotonic interactions, failing to capture complex patterns like U-shaped relationships where the influence of one concept on another may change direction \cite{quasi_nonlinear_limit_2023}. A critical limitation of standard FCMs is their reliance on static scalar weights, which forces causal relationships to be strictly monotonic. A positive weight $w_{ij} > 0$ implies that an increase in concept $c_j$ always promotes $c_i$, regardless of the current level of $c_j$. However, complex real-world systems frequently exhibit non-monotonic dynamics where the direction of influence (from promotional to inhibitory) changes based on the state of the causing concept. 

\begin{figure}[!t]
\centering
\begin{tikzpicture}
\begin{axis}[
    width=8cm, height=6cm,
    axis lines=left,
    xlabel={Concept state $c_j(t)$},
    ylabel={Influence $\phi_{ij}$},
    legend style={
        font=\footnotesize,       
        at={(0.5,-0.25)},          
        anchor=north,              
        legend columns=1,          
        cells={anchor=west},       
        fill=none,                 
        draw=none                  
    },
    ymin=-0.2, ymax=1.0,
    xmin=0, xmax=1.1,
    grid=major,
    grid style={dashed, gray!30}
]

\addplot [blue, ultra thick, smooth] {-4*(x-0.5)^2 + 0.8};
\addlegendentry{KA-FCM (non-monotonic)}

\addplot [red, dashed, thick] {0.6*x};
\addlegendentry{Standard FCM (monotonic)}

\draw[-{Stealth[scale=1.2]}, black, thick] (axis cs:0.05, 0.1) -- (axis cs:0.35, 0.65) 
    node[midway, above, sloped] {\footnotesize Positive Influence};

\draw[fill=blue] (axis cs:0.5, 0.8) circle (2.5pt);
\node[above, blue] at (axis cs:0.5, 0.82) {\footnotesize Turning Point};

\draw[-{Stealth[scale=1.2]}, black, thick] (axis cs:0.65, 0.65) -- (axis cs:0.95, 0.1) 
    node[midway, above, sloped] {\footnotesize Negative Influence};

\end{axis}
\end{tikzpicture}
\caption{Causal relationship comparison}
\label{fig:ushaped_edge}
\end{figure}
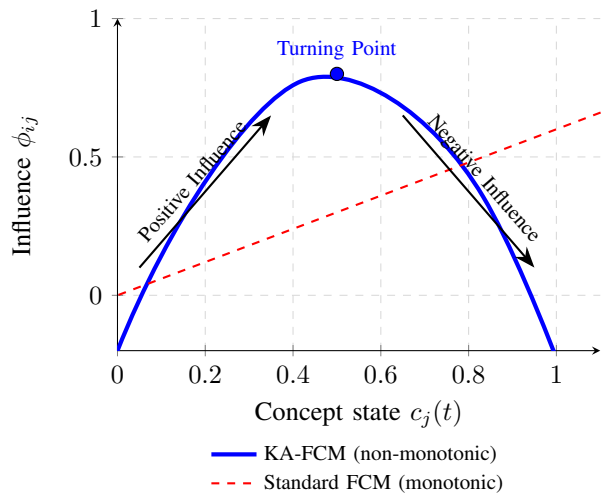

A classic example is the inverted U-shaped relationship, such as the effect of workload on productivity: initial increases in workload improve productivity up to an optimal point, beyond which further increases lead to saturation and a decrease in productivity. As shown in Figure \ref{fig:ushaped_edge}, the proposed approach overcomes this limitation. By replacing the static scalar weight with a learnable univariate function (which we will denote as $\phi_{ij}$), the model can capture a changing derivative across the input domain, effectively representing transitions where the causal influence shifts from positive to negative.

\subsection{Kolmogorov-Arnold Fuzzy Cognitive Maps}

The proposed architecture is built upon the Kolmogorov-Arnold representation theorem \cite{kolmogorov1957representation}, which establishes that any multivariate continuous function can be represented as a finite composition of univariate continuous functions and addition. Formally:

\begin{equation}
    f(x_1, \dots, x_n) = \sum_{q=0}^{2n} \Phi_q \left( \sum_{p=1}^{n} \phi_{qp}(x_p) \right)
\end{equation}

\begin{figure}[!t]
    \centering
    \includegraphics[width=6.5cm]{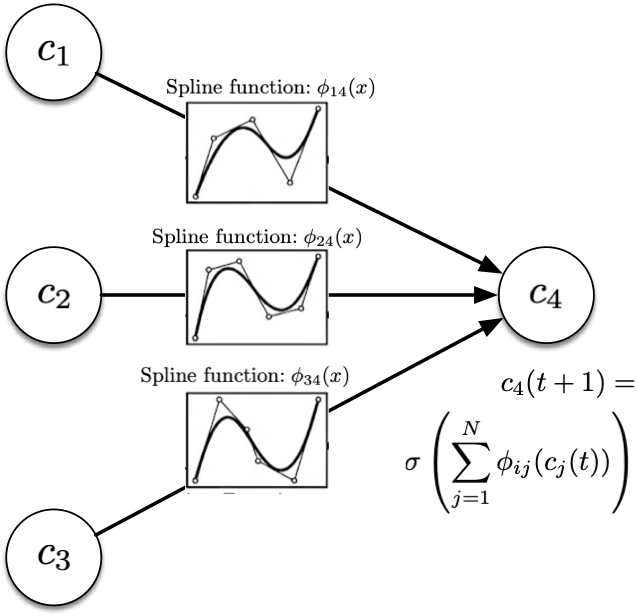}
    \caption{Kolmogorov-Arnold Fuzzy Cognitive Map architecture.}
    \label{fig:kafcm_architecture}
\end{figure}

In the context of the proposed Kolmogorov-Arnold Fuzzy Cognitive Maps, this theorem implies that causal interactions can be decomposed into a superposition of learnable univariate functions $\phi_{ij}(\cdot)$. Unlike conventional FCMs that place non-linear activations at the nodes, KA-FCM places them on the edges \cite{TOMILO2025112004}, replacing the static scalar $w_{ij}$ with a learnable univariate function $\phi_{ij}: \mathbb{R} \to \mathbb{R}$ parameterized via B-splines. This approach generalizes the conventional adjacency matrix to a functional matrix $\Phi(x)$ (Equation \ref{eq:adjacency-matrix}), where each entry $\phi_{ij}(x)$ represents the specific pairwise causal profile between nodes $j$ and $i$.

\begin{equation}
    \Phi(x) = \begin{bmatrix} 
    \phi_{11}(x) & \cdots & \phi_{1N}(x) \\ 
    \vdots & \ddots & \vdots \\ 
    \phi_{N1}(x) & \cdots & \phi_{NN}(x) 
    \end{bmatrix}
    \label{eq:adjacency-matrix}
\end{equation}

As shown in Figure \ref{fig:kafcm_architecture}, the KA-FCM update rule is computed as follows
\begin{equation}
    c_i(t+1) = \sigma \left( \sum_{j=1}^N \phi_{ij}(c_j(t)) \right)
    \label{eq:kafcm_inference}
\end{equation}

\noindent where $\sigma(\cdot)$ serves as a bounding operator to keep activations within the $[0, 1]$ range. The bounding operator $\sigma(\cdot)$ serves as a non-linear activation function designed to constrain the aggregate nodal influence within the normalized fuzzy interval $[0, 1]$. Formally, let $S_i(t) = \sum_{j=1}^{N} \phi_{ij}(c_j(t))$ denote the total functional input received by node $i$. The operator $\sigma(S_i(t))$ ensures that the updated state $c_i(t+1)$ remains within the valid state space constraints.

The bounding operator is conceptually analogous to a hard-clipping function defined as $\sigma_{\text{hard}}(x) = \min(\max(0, x), 1)$. However, the computational implementation applies a continuously differentiable sigmoid function. This smooth approximation is critical to preserve gradient flow, allowing the optimization of the edge functional parameters via backpropagation.

\subsection{Edge parameterization via B-Splines}

To ensure stable optimization and avoid vanishing gradients, each edge function $\phi_{ij}(c_j(t))$ is decomposed into a global residual base function $b(x)$ —specifically implemented as the Sigmoid Linear Unit (SiLU), $b(x)=x/(1+e^{-x})$— and a learnable local spline component parameterized by B-splines \cite{liu2024kan, high_order_2023}. B-splines were specifically selected over other universal approximators, such as high-degree global polynomials or radial basis functions. Their fundamental advantage lies in their local support property. This characteristic prevents the severe edge oscillations typical of high-degree polynomials, a mathematical behavior known as Runge's phenomenon \cite{burden2015}. Furthermore, local support ensures that gradient updates during training only alter specific segments of the curve, thereby preserving the overall stability of the learned causal relationship \cite{deboor.2001, liu2024kan}. The total edge function is expressed as:
\begin{equation}
\phi_{ij}(c_j(t)) = w_{ij}^{\textrm{base}} \cdot b(c_j(t)) + w_{ij}^{\textrm{spline}} \cdot \sum_{k=1}^{K} \alpha_{ij}^k B_{k,p}(c_j(t))
\end{equation}
where $w_{ij}^{\textrm{base}}$ and $w_{ij}^{\textrm{spline}}$ are learnable scaling factors, $\alpha_{ij}^k$ are the learnable control coefficients of the spline, and $B_{k,p}$ are the B-spline basis functions of degree $p$. The chosen base function SiLU differs from unipolar sigmoid in two key ways. First, it acts like an identity function for positive values, which ensures that strong causal signals propagate without weakening. Second, the SiLU function is non-monotonic, exhibiting a local minimum near $-1.28$. This characteristic enables the model to capture complex inhibitory dynamics that a strictly monotonic sigmoid cannot represent.

The B-spline basis functions $B_{k,p}(c_j(t))$ are constructed recursively using the Cox-de Boor formula. For degree $p=0$:
\begin{equation}
    B_{k,0}(c_j(t)) = \begin{cases} 
    1 & \text{if } t_k \leq c_j(t) < t_{k+1} \\
    0 & \text{otherwise}
    \end{cases}
\end{equation}
where $t_k$ is the $k$-th partition point in the non-decreasing sequence $\mathcal{T} = \{t_0, t_1, \dots, t_m\}$, which serves to partition the input domain into a grid of intervals. These partition points define the local support of each basis function, ensuring that $B_{k,p}(c_j(t))$ remains non-zero only within the specific range $[t_k, t_{k+p+1})$. By determining the boundaries where different polynomial segments meet, the sequence of grid points governs the smoothness and continuity of the spline, allowing the KA-FCM to approximate complex causal dynamics with high numerical stability and local precision. For higher degrees $p \ge 1$:
\begin{equation}
    \begin{aligned}
        B_{k,p}(c_j(t)) = & \frac{c_j(t) - t_k}{t_{k+p} - t_k} B_{k,p-1}(c_j(t)) \\[0.1cm]
        & + \frac{t_{k+p+1} - c_j(t)}{t_{k+p+1} - t_{k+1}} B_{k+1,p-1}(c_j(t))
    \end{aligned}
\end{equation}

This recursive structure ensures local support and guarantees smoothness. During the training process via backpropagation, the edge function evolves from the baseline behavior to approximate complex non-linear dynamics, effectively fitting patterns such as oscillations or exponential growth.

Although the proposed architecture applies learnable B-splines typically associated with function approximation, the term Fuzzy is preserved to denote the continuous nature of the concept state space $\mathcal{C} \in [0,1]^N$. Unlike Boolean networks, the KA-FCM processes degrees of concept activation rather than binary states. In this context, the functional edges $\phi_{ij}$ serve as dynamic fuzzy relations that continuously map the degree of membership of a cause to the degree of activation of an effect, effectively generalizing the traditional fuzzy implication rules found in standard Fuzzy Cognitive Maps \cite{kosko1986fuzzy}.

\subsection{Learning algorithm and regularization}

The training objective is to minimize a composite loss function $\mathcal{L}_{\text{total}}$ that balances predictive accuracy with structural parsimony. The reconstruction loss $\mathcal{L}_{\textrm{rec}}$ quantifies the divergence between the ground truth states $c$ and the predicted states $\hat{c}$ over a discrete time trajectory of length $T$:

\begin{equation}
    \mathcal{L}_{\text{rec}} = \frac{1}{T} \sum_{t=1}^{T} d\big(c(t+1), \hat{c}(t+1)\big)
\end{equation}

To prevent overfitting and enhance interpretability, an $L_1$ penalty is applied to the spline coefficients $\alpha_{ij}^k$, inducing sparsity by driving negligible interactions toward zero. The total loss is defined as follows

\begin{equation}
    \mathcal{L}_{\text{total}} = \mathcal{L}_{\text{rec}} + \lambda \Omega(\Phi)
\end{equation}
where $\lambda$ controls the trade-off between accuracy and complexity, and the regularization term is defined as the $L_1$ norm of the spline control points: $\Omega(\Phi) = \sum_{i,j,k} |\alpha_{ij}^k|$. Parameters $\Theta = \{w^{\text{base}}, w^{\text{spline}}, \alpha\}$ are updated using gradient-based optimization and automatic differentiation. 

Given the non-convex nature of the loss landscape associated with deep neuro-symbolic architectures, strict theore\-tical guarantees of global convergence cannot be established. However, empirical observations across multiple initializations indicate that the local support of the B-splines, combined with the residual SiLU connections, consistently guides the gradient descent toward stable, high-quality local optima without suffering from severe vanishing gradients. 

The complete total loss equation is as follows:
\begin{equation}
    \mathcal{L}_{\text{total}} = \frac{1}{T} \sum_{t=1}^{T} d\big(c(t+1), \hat{c}(t+1)\big) + \lambda \sum_{i,j,k} |\alpha_{ij}^k|
\end{equation}

Regarding computational complexity, a standard FCM inference step scales as $O(N^2)$. In the proposed KA-FCM, evaluating $N^2$ spline functions over a grid of size $G$ with degree $p$ results in a complexity proportional to $O(N^2 \cdot p^2)$, leveraging the local support property of B-splines. Since $G$ and $p$ are small fixed constants (e.g., $G \approx 10, p=3$) independent of the network size, the asymptotic complexity remains quadratic $O(N^2)$. Although this introduces a constant computational overhead per edge evaluation, the asymptotic growth rate $O(N^2)$ with respect to the number of concepts remains invariant. This makes the architecture computationally tractable for standard FCM topologies while significantly expanding their representational capacity.

\section{Experimental approach}
\label{sec:exps}

The experimental approach is structured around three distinct case studies, each designed to isolate and validate specific capabilities of the proposed architecture. Firstly, \textit{Experiment I: The Non-Monotonicity Paradox} (section \ref{exp:1}) assesses the capacity of the model to capture non-linear dynamics that typically challenge traditional linear inference mechanisms. Secondly, \textit{Experiment II: Symbolic Causal Discovery} (section \ref{exp:2}) examines the symbolic regression capabilities of the framework, focusing on the interpretability and accurate recovery of underlying mathematical laws. Finally, \textit{Experiment III: Mackey-Glass Time Series Forecasting} (section \ref{exp:3}) evalu\-ates the predictive performance and robustness of the method when applied to chaotic and complex temporal systems.

To determine the optimal configuration for the proposed architecture, a hyperparameter optimization was conducted via grid search. This procedure entailed an exhaustive evaluation across a predefined search space, specifically targeting the grid resolution $G$, the learning rate $\eta$, and the training duration. The grid resolution was explored within the integer range $G \in [4, 19]$, while the learning rate was evaluated across the set $\eta \in \{0.001, 0.01, 0.05, 0.1\}$. In addition, the sensitivity to training duration was assessed by varying the number of epochs across ten linearly spaced values within the interval $[500, 1500]$. 

The selection of the best-performing model was based on minimizing the error metric on the validation set. While the theoretical framework supports $L_1$ regularization to induce sparsity, in these experiments, the regularization coefficient was set to $\lambda = 0$. This configuration was chosen to assess the maximum representational capacity of the splines and to allow the model to capture high-frequency details in the chaotic regimes without constraints.

The comparative analysis included a standard FCM utilizing a hyperbolic tangent activation function and a conventional Multi-Layer Perceptron (MLP), contrasting their performance with the best KA-FCM. 

The baseline MLP consists of three fully connected layers separated by non-linear activation functions. Let $\mathbf{x} \in \mathbb{R}^{N_{in}}$ denote the input vector. The network processes the input through two hidden layers of size $d_{\textrm{model}}=64$ and generates the output $\hat{\mathbf{y}}$ as follows

\begin{equation}
    \begin{aligned}
        \mathbf{h}_1 &= \text{ReLU}(\mathbf{W}_1 \mathbf{x} + \mathbf{b}_1) \\
        \mathbf{h}_2 &= \text{ReLU}(\mathbf{W}_2 \mathbf{h}_1 + \mathbf{b}_2) \\
        \hat{\mathbf{y}} &= \tanh(\mathbf{W}_3 \mathbf{h}_2 + \mathbf{b}_3)
    \end{aligned}
\end{equation}

\noindent where $\mathbf{W}_1 \in \mathbb{R}^{64 \times N_{\textrm{in}}}$, $\mathbf{W}_2 \in \mathbb{R}^{64 \times 64}$, and $\mathbf{W}_3 \in \mathbb{R}^{N_{\textrm{out}} \times 64}$ are the learnable weight matrices, and $\mathbf{b}_l$ are the corresponding bias vectors. The final activation function serves to constrain the system output to the interval $[-1, 1]$.

\subsection{Experiment I: Modeling non-monotonic relationships using the Yerkes-Dodson Law}
\label{exp:1}

The first experiment addresses the fundamental limitation of standard FCMs: the inability to model non-monotonic relations. This experiment employed a synthetic dataset representing the well-known Yerkes-Dodson Law from the field of psychology. This law postulates an inverted U-shaped relationship, asserting that performance improves with increasing arousal only up to an optimal threshold; beyond this point, excessive stimulation leads to a deterioration in functioning. The task was to learn this relationship from data, using the Mean Squared Error (MSE) on a hold-out test set as the performance metric.

The experimental topology consists of a simple 2-node cognitive map, with concept 1 (stress/arousal) modulating concept 2 (performance). The input variable was uniformly sampled from the normalized domain $x \in [-1, 1]$ ($n=1000$). The ground truth response was modeled using a scaled Gaussian function centered at zero to reproduce the theoretical inverted U-shape, following the equation:

\begin{equation}
    y = 1.6 \exp(-4x^2) - 1 + \epsilon
\end{equation}

\noindent where the term $1.6 \exp(-4x^2) - 1$ maps the peak performance to $y=0.6$ (at optimal stress $x=0$) and degrades performance towards $y \approx -1$ at the extremes ($x = \pm 1$). To evaluate model robustness against stochastic variability, additive Gaussian noise $\epsilon \sim \mathcal{N}(0, 0.05^2)$ was injected into the output signal.

\begin{figure}[!t]
\centering
\includegraphics[width=\columnwidth]{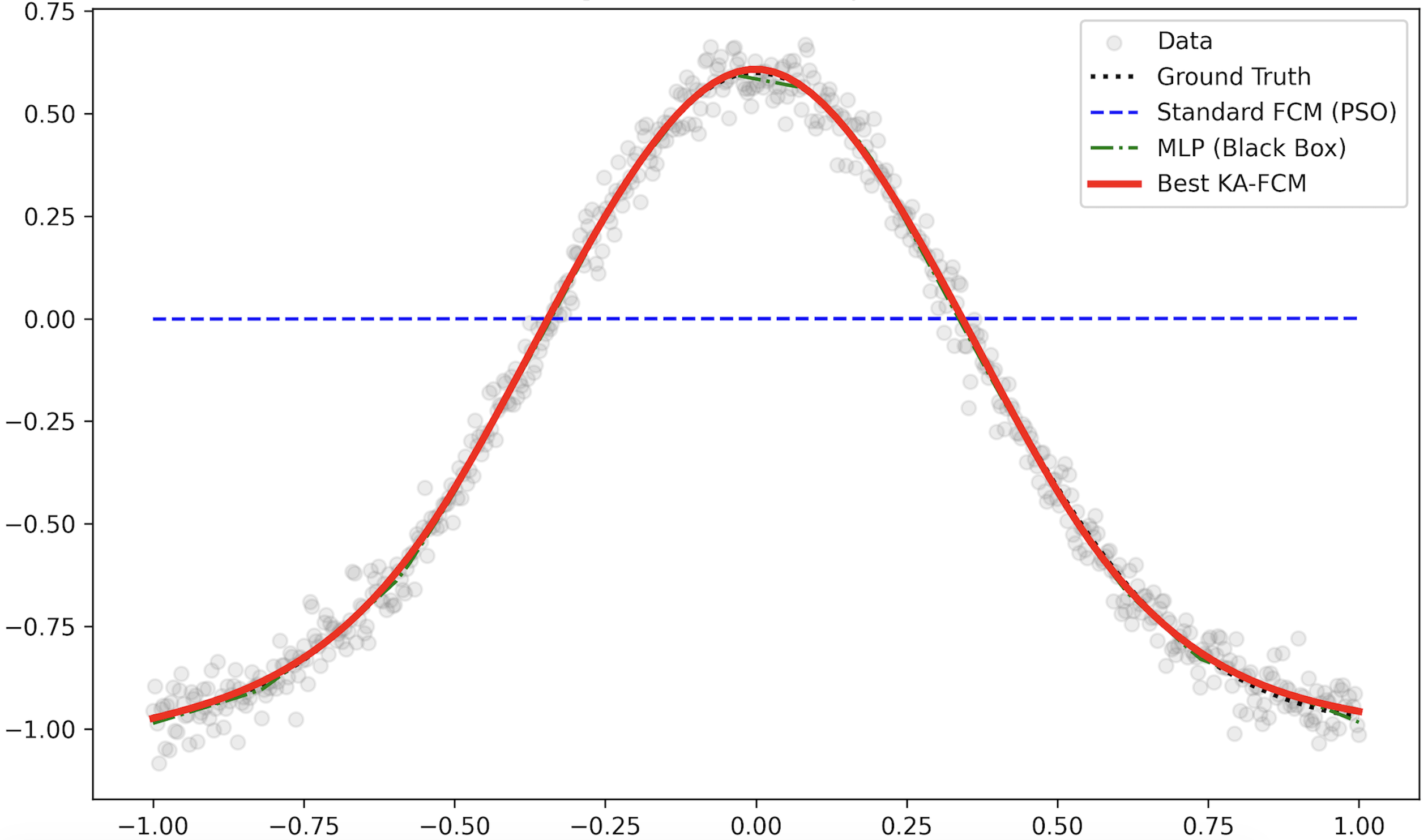}
\caption{Comparative modeling of the non-monotonic Yerkes-Dodson law. The scatter plot represents the synthetic ground truth exhibiting an inverted U-shape. The dashed blue line indicates the standard FCM prediction. The green dashed line shows the MLP results. The solid red line shows the function learned by the KA-FCM edge.}
\label{fig:yerkes}
\end{figure}

Fig. \ref{fig:yerkes} illustrates the comparative modeling capabilities. The standard FCM fails to represent the dual behavior (positive and negative slope) within a single edge, effectively predicting a flat line (dashed blue). The best performing KA-FCM model was obtained with a grid size of $G=4$, a learning rate of $\eta=0.1$, and $610$ training epochs. The red curve shows that the KA-FCM successfully captures both the amplification and inhibition phases without the need for hidden layers. Similarly, the MLP baseline accurately approximates the underlying non-linear function.

The results (Table \ref{tab:results-exp1}) demonstrate a significant difference in performance between the architectures. The standard FCM failed to capture the underlying dynamic, leading to a high MSE. During training, the scalar weight $w_{12}$ converged to a value near zero because the linear correlation between the input $x$ and the output $y$ in a symmetric bell curve is approximately zero. This results in a constant mean prediction, explaining the high error. While the MLP baseline successfully approximated the non-linear function, the KA-FCM achieved the lowest MSE, confirming its ability to model complex saturation and non-monotonic dynamics that are structurally impossible for standard causal maps.

Furthermore, the stability analysis in Table \ref{tab:results-exp1} highlights the robustness of the proposed method. The KA-FCM achieved the lowest maximum absolute error ($0.015$) and standard deviation ($0.006$), outperforming the MLP baseline ($0.025$ and $0.009$, respectively).  This indicates that the spline-based edge not only fits the global trend but also maintains tighter control over local deviations compared to the black-box neural network.

\begin{table}[h]
    \centering
    \caption{Performance and stability analysis (Experiment I)}
    \label{tab:results-exp1}
    \begin{tabular}{l c c c}
        \hline
        \textbf{Model} & \textbf{MSE} & \textbf{Max. Abs. Error} & \textbf{Std. Dev. Error} \\
        \hline
        Standard FCM & $3.96 \times 10^{-1}$ & $0.971$ & $0.556$ \\
        MLP & $1.18 \times 10^{-4}$ & $0.025$ & $0.009$ \\
        \textbf{KA-FCM} & $\mathbf{4.10 \times 10^{-5}}$ & $\mathbf{0.015}$ & $\mathbf{0.006}$ \\
        \hline
    \end{tabular}
\end{table}

Table \ref{tab:exp1-hyperparams}  shows a sensitivity analysis of the hyperparameters. It reveals that the learning rate is the dominant factor governing convergence, exhibiting a strong negative correlation with the error ($\rho \approx -0.61$). It is important to highlight that the grid size displayed an insignificant correlation with the error ($+0.05$). This observation implies that for smooth, unimodal topologies such as the inverted U-shape, a sparse grid ($G=4$) affords enough representational capacity without compromising generalization.

\begin{table}[h]
    \centering
    \caption{Hyperparameter sensitivity (Experiment I)}
    \label{tab:exp1-hyperparams}
    \begin{tabular}{lcc}
        \hline
        \textbf{Parameter} & \textbf{Optimal value} & \textbf{Correlation w/ Error} \\
        \hline
        Grid size ($G$) & $4$ & $+0.05$ (minimal) \\
        Learning rate ($\eta$) & $0.1$ & $\mathbf{-0.61}$ (critical) \\
        Epochs & $611$ & $-0.16$ (low) \\
        \hline
    \end{tabular}
\end{table}

\subsection{Experiment II: Symbolic causal discovery}
\label{exp:2}

The second experiment explores the capacity of KA-FCM to act as a tool for scientific discovery. In many scientific fields, the goal is not just to predict, but to uncover the mathematical laws governing the system. A synthetic dataset was constructed to represent a relationship where concept B exhibits a sinusoidal dependency on concept A, defined by $y = \sin(3x)$. 

Conventional causal weights are structurally insufficient for modeling such periodicity. As linear operators, they allow only for magnitude scaling and cannot approximate non-monotonic functions. Following the training of a KA-FCM on this data, the learned edge function was analyzed. To validate the interpretability of the model, symbolic regression was applied to the approximated spline curve to determine whether the original sinusoidal function could be accurately recovered from the learned parameters.

\begin{figure}[!t]
\centering
\includegraphics[width=\columnwidth]{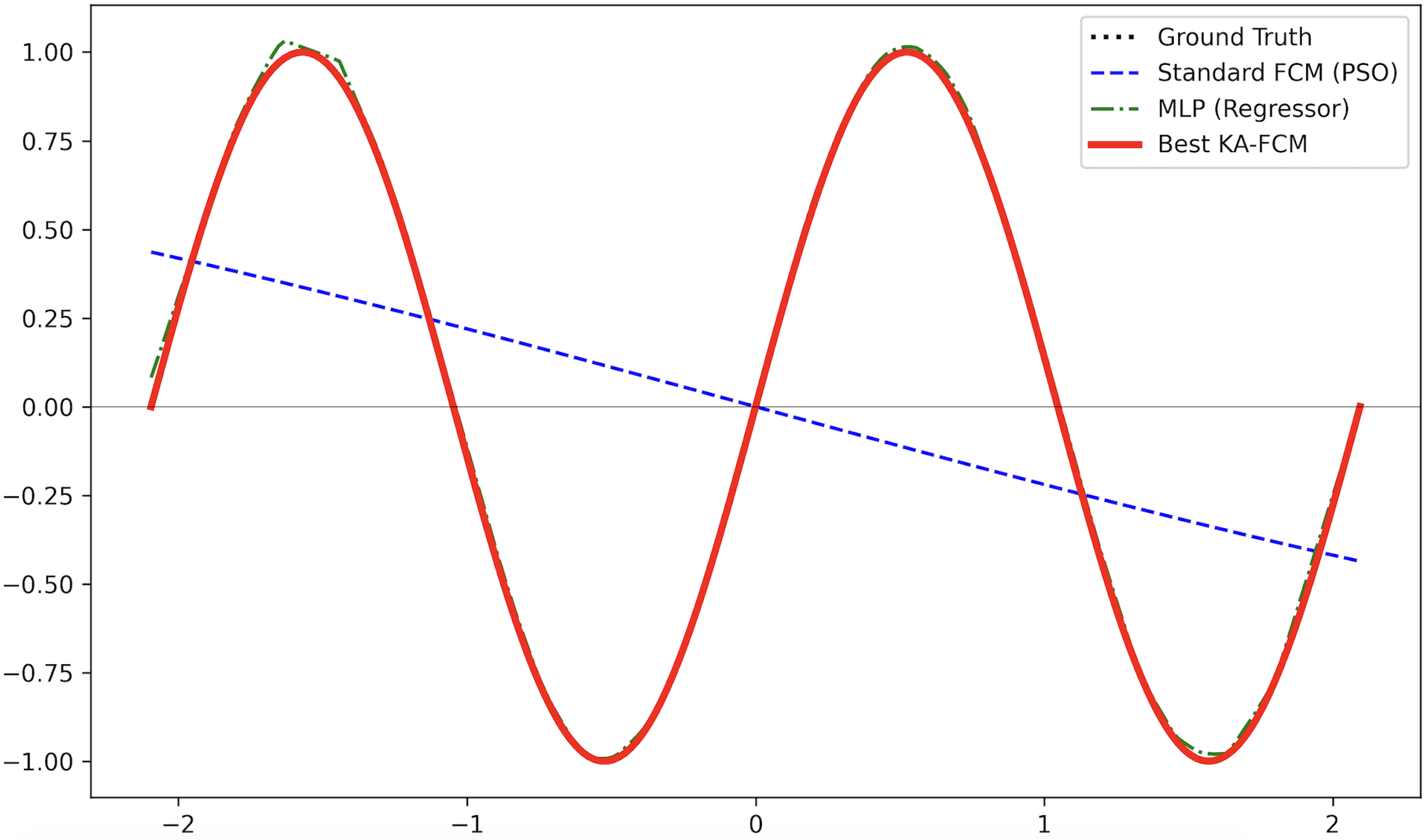}
\caption{Learned edge functions comparison.}
\label{fig:symbolic}
\end{figure}

The experimental results are shown in Table \ref{tab:results-exp2} and Figure \ref{fig:symbolic}. The standard FCM, restricted by its monotonic activation constraints, failed to capture the periodic oscillations, converging instead to a linear approximation that results in significant information loss. The Multi-Layer Perceptron successfully approximated the non-linear dynamics, achieving a high-fidelity fit comparable to the ground truth. However, the quantitative assessment in Table \ref{tab:results-exp2} highlights a critical distinction between approximation and exact recovery. While the MLP achieved a low MSE ($4.37 \times 10^{-4}$), the KA-FCM attained an error four orders of magnitude lower ($3.40 \times 10^{-8}$). Moreover, the stability metrics reveal that the KA-FCM's maximum absolute error was merely $0.0004$, compared to $0.082$ for the MLP. 

\begin{table}[h]
    \centering
    \caption{Performance and stability analysis (Experiment II)}
    \label{tab:results-exp2}
    \begin{tabular}{l c c c}
        \hline
        \textbf{Model} & \textbf{MSE} & \textbf{Max. Abs. Error} & \textbf{Std. Dev. Error} \\
        \hline
        Standard FCM & $4.26 \times 10^{-1}$ & $1.119$ & $0.654$ \\
        MLP (Baseline) & $4.37 \times 10^{-4}$ & $0.082$ & $0.011$ \\
        \textbf{KA-FCM} & $\mathbf{3.40 \times 10^{-8}}$ & $\mathbf{0.0004}$ & $\mathbf{0.0002}$ \\
        \hline
    \end{tabular}
\end{table}

The KA-FCM addresses the MLP's lack of interpretability through its transparent structure. Rather than simply memorizing the training points, the function captured the underlying sinusoidal pattern. Furthermore, the interpretable nature of the spline control points enabled the application of symbolic regression, allowing the recovery of an analytical expression closely approximating $y = \sin(3x)$. These results confirm that the KA-FCM architecture successfully combines the high accuracy of deep neural networks with the transparency of FCMs, effectively bridging the gap between data-driven learning and theoretical modeling.

The hyperparameter analysis in Table \ref{tab:exp2-hyperparams} indicates that the grid size ($G$) here showed a significant negative correlation with the error ($\rho = -0.32$). The optimal configuration required a denser grid ($G=19$) to accurately resolve the inflection points of the sine wave. The learning rate remained a high-impact factor ($\rho = -0.40$), reinforcing the need for aggressive updates to escape local minima in oscillatory loss landscapes.

\begin{table}[h]
    \centering
    \caption{Hyperparameter sensitivity (Experiment II)}
    \label{tab:exp2-hyperparams}
    \begin{tabular}{l c c}
        \hline
        \textbf{Parameter} & \textbf{Optimal value} & \textbf{Correlation w/ Error} \\
        \hline
        Grid size ($G$) & $19$ & $\mathbf{-0.32}$ (significant) \\
        Learning rate ($\eta$) & $0.1$ & $\mathbf{-0.40}$ (high) \\
        Epochs & $1500$ & $-0.22$ (moderate) \\
        \hline
    \end{tabular}
\end{table}

\subsection{Experiment III: Mackey-Glass chaotic time series forecasting}
\label{exp:3}

This experiment evaluates the model's performance on a standard benchmark for chaotic systems: the Mackey-Glass differential equation (Equation \ref{eq:mackey-glass}). 
\begin{equation}
    \frac{dx(t)}{dt} = \beta \frac{x(t-\tau)}{1 + x(t-\tau)^n} - \gamma x(t)
    \label{eq:mackey-glass}
\end{equation}

For this experiment, the parameters were fixed at $\beta=0.2$ (production rate), $\gamma=0.1$ (decay constant), and $n=10$ (non-linearity factor). The time delay parameter was set to $\tau=17$, a value known to induce chaotic dynamics.

This series exhibits chaotic behavior that is notoriously difficult to predict with linear models due to its sensitivity to initial conditions. For the modeling phase, a lag ($lag=4$) embedding approach was adopted, utilizing $x(t-4)$, $x(t-3)$, $x(t-2)$, and $x(t-1)$ as predictor nodes for the target $x(t)$. The selected metric adopted for evaluation is the Mean Absolute Percentage Error (MAPE).
\begin{equation}
    \text{MAPE} = \frac{100\%}{T} \sum_{t=1}^{T} \left| \frac{c(t) - \hat{c}(t)}{c(t)} \right|
\end{equation}

\begin{figure}[!t]
\centering
\includegraphics[width=\columnwidth]{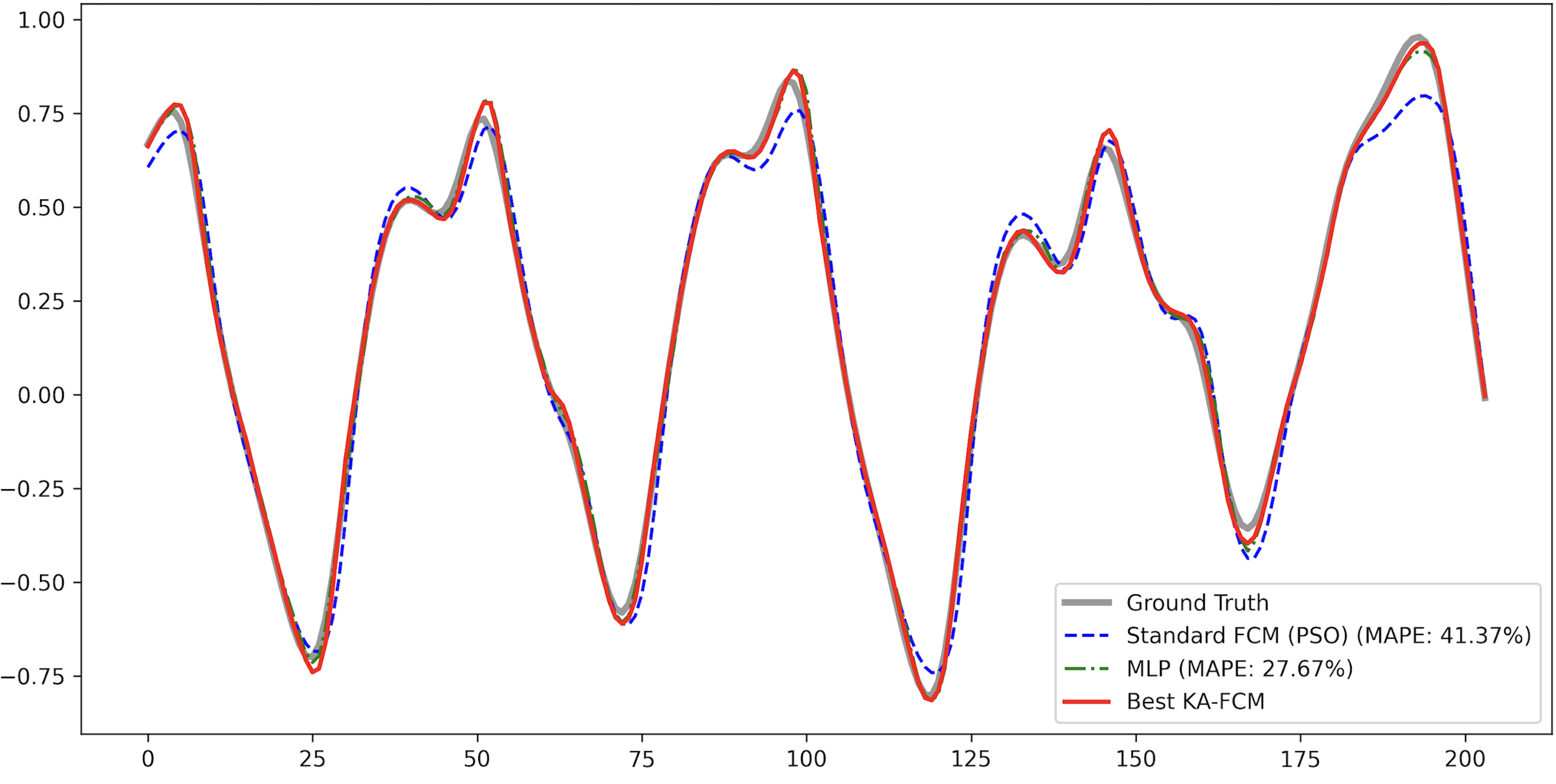}
\caption{One-step-ahead forecasting performance on the chaotic Mackey-Glass time series ($\tau=17$).}
\label{fig:mackey}
\end{figure}

The experimental results presented in Table \ref{tab:results-exp3} and Figure \ref{fig:mackey} demonstrate that the proposed method outperforms the baseline models. The standard FCM resulted in the highest error rate ($41.37\%$), as it proved unable to follow the high-frequency dynamics of the chaotic attractor; its linear aggregation mecha\-nism smoothed out the chaotic details, resulting in a poor fit. The MLP performed significantly better, capturing the general temporal dynamics. However, the KA-FCM outperformed both baselines by a substantial margin.

Quantitatively, the KA-FCM achieved a MAPE of $\mathbf{13.66\%}$, effectively halving the error rate of the MLP baseline ($27.67\%$). Furthermore, the stability metrics indicate that the proposed model produces more consistent predictions, with the lowest standard deviation of errors ($0.023$) and the tightest bound on maximum absolute error ($0.072$). The flexibility of the edges to model arbitrary non-linear functions enables the KA-FCM to reconstruct the chaotic phase space with superior fidelity compared to the alternatives. 

\begin{table}[h]
    \centering
    \caption{Performance and stability analysis (Experiment III - Mackey-Glass)}
    \label{tab:results-exp3}
    \begin{tabular}{l c c c}
        \hline
        \textbf{Model} & \textbf{MAPE (\%)} & \textbf{Max. Abs. Error} & \textbf{Std. Dev. Error} \\
        \hline
        Standard FCM & $41.37\%$ & $0.162$ & $0.058$ \\
        MLP & $27.67\%$ & $0.097$ & $0.027$ \\
        \textbf{KA-FCM} & $\mathbf{13.66\%}$ & $\mathbf{0.072}$ & $\mathbf{0.023}$ \\
        \hline
    \end{tabular}
\end{table}

The hyperparameter sensitivity analysis (Table \ref{tab:exp3-hyperparams}) details distinct optimization dynamics compared to the previous experiments. The learning rate remains the critical determinant of performance, showing a strong negative correlation with the error ($\rho = -0.62$); a moderate rate of $\eta=0.05$ proved optimal for balancing convergence speed with stability in the chaotic landscape. Interestingly, while the correlation between grid size and error was slight ($-0.07$), the optimal configuration utilized a dense grid ($G=19$). This suggests that while increasing grid resolution provides the potential for better fitting, the optimization is less sensitive to this parameter than to the step size of the gradient descent.

\begin{table}[h]
    \centering
    \caption{Hyperparameter sensitivity (Experiment III)}
    \label{tab:exp3-hyperparams}
    \begin{tabular}{l c c}
        \hline
        \textbf{Parameter} & \textbf{Optimal value} & \textbf{Correlation w/ Error (MAPE)} \\
        \hline
        Grid size ($G$) & $19$ & $-0.07$ (slight) \\
        Learning rate ($\eta$) & $0.05$ & $\mathbf{-0.62}$ (critical) \\
        Epochs & $1277$ & $-0.28$ (moderate) \\
        \hline
    \end{tabular}
\end{table}

\subsection{Computational cost analysis}

To provide an evaluation of the proposal, it is necessary to analyze the computational cost associated with the KA-FCM compared to the standard FCM and the MLP baselines across the experimental scenarios. Because wall-clock training times are heavily dependent on hardware configurations and the underlying software implementation (e.g., parallel tensor processing vs. sequential execution), we assess the computational cost in terms of structural complexity and relative operational overhead.

The standard FCM exhibits the lowest computational cost. Its architecture relies on a static adjacency matrix, requiring only $O(N^2)$ scalar parameters and simple matrix multiplications during both training and inference.

In contrast, the proposed KA-FCM replaces these scalar weights with learnable univariate functions. For a grid size $G$, the number of trainable control coefficients scales as $O(N^2 \cdot G)$. Consequently, during the training phase, the evaluation of the B-spline basis functions, the computation of the SiLU residuals, and the subsequent backpropagation through the control grid introduce a significant computational overhead per epoch compared to the standard FCM. The training cost of the KA-FCM is generally comparable to that of the MLP baseline, as both models require gradient descent over a complex, non-convex loss landscape to approximate non-linear dynamics.

However, it is crucial to distinguish between training and inference costs. While the training phase of the KA-FCM is more computationally intensive, the inference phase remains highly efficient. Once the optimal B-spline control points ($\alpha_{ij}^k$) are learned, a forward pass through the network requires only the evaluation of localized polynomial segments. Because the asymptotic complexity of the inference step remains bounded by $O(N^2)$, the execution time post-training is extremely fast. This ensures that, despite the heavier training process, the KA-FCM is fully viable for real-time prediction and control applications.

\section{Conclusion}
\label{sec:conclusion}

This research presents the Kolmogorov-Arnold Fuzzy Cognitive Map (KA-FCM), a neuro-symbolic architecture that evolves the traditional causal modeling framework from static scalar weights to dynamic learnable functions. By replacing the constant coefficients $w_{ij}$ with univariate spline-based networks, the proposed model successfully addresses the structural inability of standard FCMs to represent non-monotonic and complex non-linear relationships.

The experimental validation provides empirical evidence of the model's capabilities across three distinct problem domains. First, in modeling the Yerkes-Dodson law, the KA-FCM demonstrated that it can capture inverted U-shaped relationships that are strictly impossible for linear aggregators. Second, the symbolic discovery experiment confirmed that the architecture does not merely approximate the data but can accurately recover the underlying analytical law (e.g., $y=\sin(3x)$) with an error magnitude of $10^{-8}$, significantly outperforming the Multi-Layer Perceptron results. Third, in the chaotic Mackey-Glass forecasting task, the method achieved superior stability and reduced the Mean Absolute Percentage Error (MAPE) by approximately 50\% compared to the MLP, proving its robustness in tracking high-frequency dynamics.

Across the three experiments, a generalizable heuristic for hyperparameter tuning emerges: the optimal grid size is heavily dependent on the frequency of the underlying dynamic. Smooth, low-frequency causal relationships (e.g., Yerkes-Dodson) are adequately modeled with sparse grids, preventing overfitting. Conversely, highly non-linear or chaotic regimes (e.g., Mackey-Glass) demand denser grids to capture rapid inflection points. In all cases, the learning rate proved to be the most critical parameter, requiring careful tuning to navigate the non-convex optimization landscape of the splines.

A key strength of this proposal is that it balances predictive accuracy with a higher degree of structural transparency than fully connected deep neural networks. While we acknowledge that analyzing a dense functional adjacency matrix in large-scale systems remains challenging, the KA-FCM provides explicit functional forms for each specific edge. As demonstrated in the symbolic regression task, this allows for the direct inspection and extraction of pairwise causal relationships without the obfuscation introduced by hidden node layers. This is an essential feature for deployment in safety-critical domains such as medicine and engineering.

However, this approach comes with computational costs. Replacing scalar weights with functions increases the number of parameters per edge, leading to a memory footprint that scales with the grid size ($G$). While the computational overhead of spline evaluation is manageable for typical topologies, regularization strategies are essential to ensure the physical plausibility of the learned functions and to prevent overfitting in sparse data regimes.


\bibliographystyle{IEEEtran}
\bibliography{refs}

\section{Biography}
\begin{IEEEbiography}[{\includegraphics[width=1in,height=1.25in,clip,keepaspectratio]{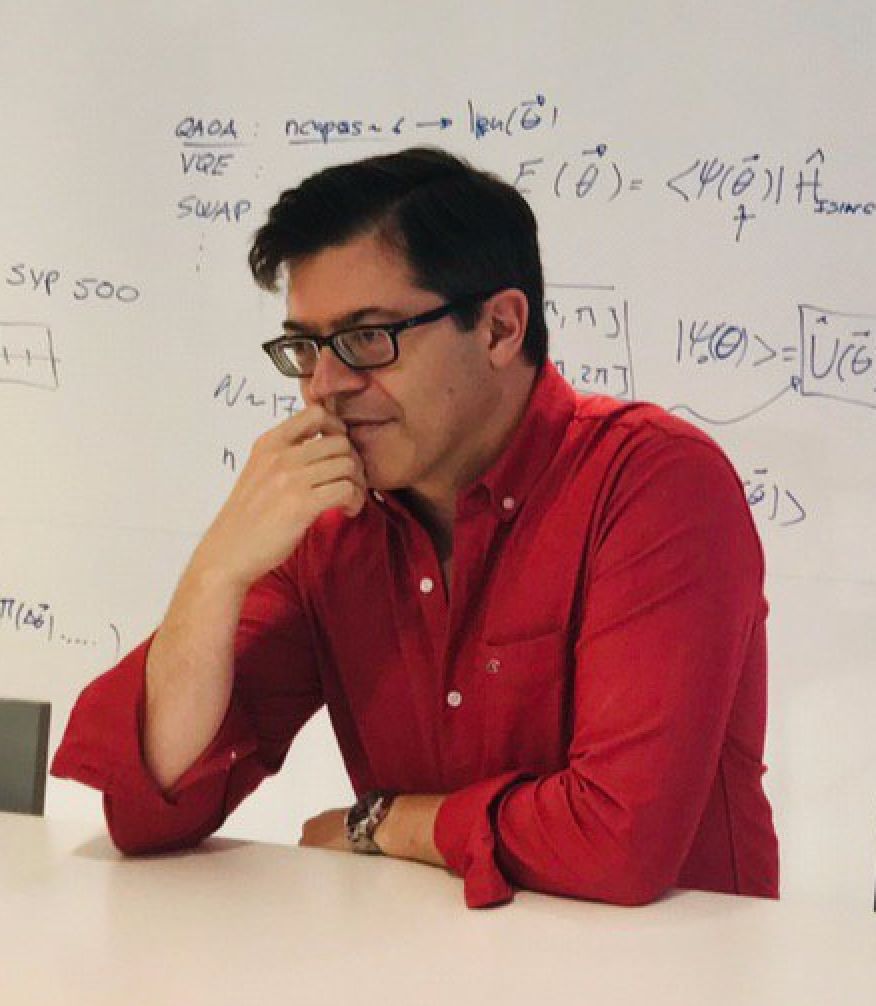}}]
{Jose L. Salmeron}~is a Professor in Artificial Intelligence with CUNEF University. He has almost 30 years of experience in technology and research, including academic positions at several universities, consulting in the IT industry, and a wide range of collaborations with private and public organizations. Moreover, he has been actively involved (as leader and team member) in EU and professional projects, working with the development of intelligent algorithms for decision support, Fuzzy Cognitive Maps and new methodologies based on soft computing, artificial intelligence techniques for complex diagnostic, decision support, and quantitative methods. 
His research has been published in top-tier journals such as IEEE Transactions on Cybernetics, IEEE Transactions on Fuzzy Systems, IEEE Transactions on Software Engineering, Expert Systems with Applications, Communications of the ACM, Journal of Systems and Software, Future Generation Computer Systems, Applied Soft Computing, Engineering Applications of Artificial Intelligence, Neurocomputing, and Information Sciences, among others. In addition, he is recognized as an ACM lifetime senior member. Currently, his research interests include privacy-preserving computing, distributed artificial intelligence, explainable artificial intelligence, and quantum machine learning.
\end{IEEEbiography}

\vfill

\end{document}